\documentclass[10pt,twocolumn,letterpaper]{article}

\usepackage{cvpr}
\usepackage{times}
\usepackage{epsfig}
\usepackage{graphicx}
\usepackage{amsmath}
\usepackage{amssymb}

\usepackage[pagebackref=true,breaklinks=true,letterpaper=true,colorlinks,bookmarks=false]{hyperref}

\cvprfinalcopy 

\def\cvprPaperID{8755} 

\ifcvprfinal\fi
\usepackage{algorithm, algorithmic}
\usepackage{multirow}
\usepackage{subfig}
\usepackage{url}
\usepackage[all]{nowidow}

\begin{document}

\title{It GAN DO Better: GAN-based Detection of Objects on Images\\ with Varying Quality}

\author{Charan D. Prakash\\
Arizona State University\\
{\tt\small cprakash@asu.edu}
\and
Lina J. Karam\\
Arizona State University\\
{\tt\small karam@asu.edu}
}

\maketitle
\begin{abstract}
    In this paper, we propose in our novel generative framework the use of Generative Adversarial Networks (GANs) to generate features that provide robustness for object detection on reduced quality images. The proposed GAN-based Detection of Objects (GAN-DO) framework is not restricted to any particular architecture and can be generalized to several deep neural network (DNN) based architectures. The resulting deep neural network maintains the exact architecture as the selected baseline model without adding to the model parameter complexity or inference speed. We first evaluate the effect of image quality not only on the object classification but also on the object bounding box regression. We then test the models resulting from our proposed GAN-DO framework, using two state-of-the-art object detection architectures as the baseline models. We also evaluate the effect of the number of re-trained parameters in the generator of GAN-DO on the accuracy of the final trained model. Performance results provided using GAN-DO on object detection datasets establish an improved robustness to varying image quality and a higher mAP compared to the existing approaches.
\end{abstract}


\section{Introduction}

\begin{figure}[!t]
  \centering
  \includegraphics[width=\linewidth,height=0.29\textheight]{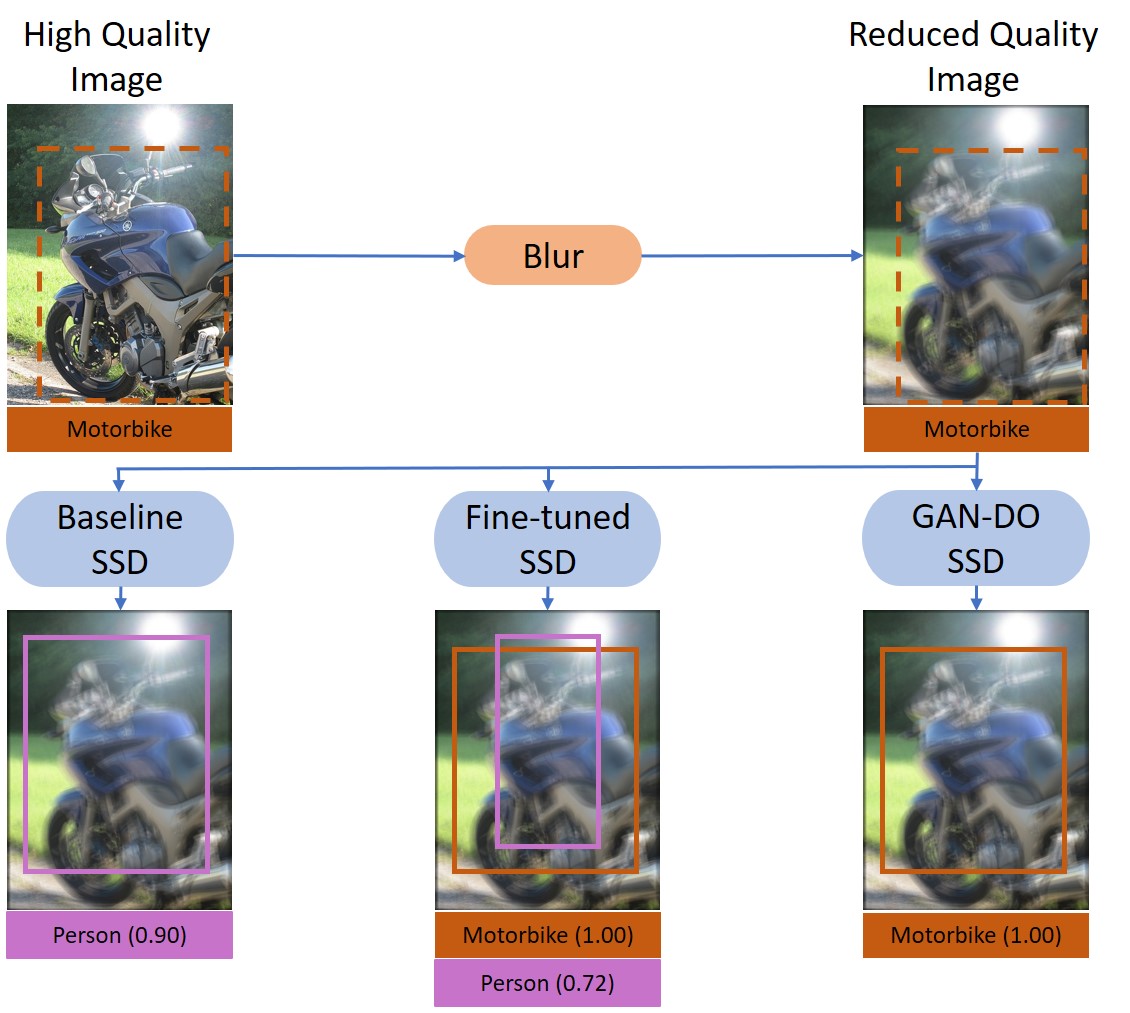}
  \caption{A reduced quality image with ground-truth object shown using a dotted bounding box is used as an input for object detection for different SSD300 \cite{SSD} based DNN models: baseline SSD300 model trained on high quality images, fine-tuned SSD300 model trained on images with varying quality, and SSD300 model trained on images with varying quality using our proposed GAN-DO framework. Each text box below an image specifies the class of the corresponding object bounding box along with the predicted class confidence score.}
  \label{IntroFig_GANDO}
\end{figure}

Deep learning and neural networks have been a popular choice for computer vision based applications such as multi-class object detection and classification. The parameters \boldmath{$\theta$} of a neural network that is designed for object detection are learned by training the network to fit a function \(f(x; \theta)=y\) where \(x\) represents an image in the training dataset \(X\) and \(y \in Y\) represents the output bounding boxes and their corresponding class labels for the image \(x\). The assumption in the above training approach is that the images in the test dataset are also drawn from the same distribution as that of the training dataset, \(X\). However, in practice, the test data might not lie in the distribution \(X\), due to various factors such as defocus and camera shake blur or images affected by noise during sensor acquisition and compression. In such cases, the neural network trained on the distribution \(X\) leads to errors in object detection. Such errors in object detection could lead to several issues ranging from social controversies (e.g., person misclassified as animal) to fatal accidents (e.g., autonomous vehicles failing to detect pedestrians).  This behaviour is of concern in the light of recent studies showing that variations in image quality deteriorates the performance of DNNs significantly \cite{dodgeUnderstanding, rodnerQuality}.\par


In this paper, we propose a framework for GAN-based detection of objects (GAN-DO) that learns an adversarial objective. The proposed GAN-DO framework guides the learning in a direction such as to maximize the similarity between the features that are computed by the GAN's generator for the reduced quality images and the features that are computed by the baseline model for the original high quality images. This ability of the GAN-DO framework leads to robust object detection, as shown in Fig.~\ref{IntroFig_GANDO}. Our proposed GAN-DO framework is not restricted to a particular architecture and is designed to accommodate the use of several DNN based object detection models and to improve the accuracy and the robustness of the selected baseline model to images with varying quality. We believe this to be the first work to adopt a GAN framework for training object detection models in order to obtain robust object detection on reduced quality images.\unboldmath\par

Three contributions are made in this paper. Firstly, we evaluate the effect of varying image quality on the object classification and the object bounding box regression of object detection models. Secondly, we propose a novel generative GAN-DO framework that consists of two neural networks collectively working as a GAN that learns an adversarial objective through training, to add robustness to the object detection model. We show that the proposed GAN-DO framework outperforms the widely used fine-tuning framework in improving the accuracy and the robustness of the baseline object detection model to varying image quality while maintaining the identical model architecture, complexity and inference speed of the baseline model. Finally, we investigate the effect of the number of re-trained parameters using the proposed GAN-DO framework, on the object detection accuracy.

\section{Related Work}\label{RelatedWork}

Neural networks are known to be susceptible to variations in image quality for the task of image classification. Recent work has shown that the performance of neural networks decreases on reduced quality images \cite{dodgeUnderstanding, FaceImageQuality, rodnerQuality, BlurFinetuning}. Tests \cite{dodgeUnderstanding, FaceImageQuality} showed that some architectures such as VGG16 \cite{VGG} were more robust to variations in image quality than other architectures, such as GoogleNet \cite{GoogleNet}. Prior work mainly concentrated on the performance of DNNs for image classification and not for object detection.\par

One of the well-known approaches of improving the accuracy of the pre-trained models is to ``fine-tune'' the pre-trained model on reduced quality images. Previous work showed that fine-tuning can improve the accuracy for the task of image classification on images with distortions such as noise and blur \cite{deepcorrect, BlurFinetuning}. Vasiljevic \etal also showed that fine-tuning on a uniform mix of sharp and blurred images produced improved accuracy \cite{BlurFinetuning}.\par

In the literature, an additional constraint has been previously imposed during fine-tuning in the form of stability loss \cite{StabilityTraining} for achieving robustness in applications such as image classification and similar-image ranking. However, recent work \cite{FeatureQuantization} has shown that, while stability training works well for some types of distortions such as JPEG bit-rate compression, the performance degrades severely for other types of distortions such as noise and blur \cite{deepcorrect}. This characteristic limits the abilities of stability training since existing networks are already known to be robust to JPEG distortions \cite{dodgeUnderstanding}. Furthermore, manually choosing or designing effective stability loss terms for each distortion and task is challenging. For example, the authors of \cite{StabilityTraining} propose K-L divergence as the stability term for the task of image classification and \(L_2\) distance as the stability term for the task of feature embedding and similar-image ranking. No loss term is proposed for the task of object detection that includes bounding box regression.\par 

Feature quantization is another approach for image classification that was proposed to improve the robustness of DNNs to varying image quality without changing the DNN model complexity. Sun \textit{et al.} propose different types of additional non-linearities such as flooring and exponential power operations on the features for feature quantization \cite{FeatureQuantization}. However, tests in \cite{FeatureQuantization} show that there is no single non-linearity that performs better on all types of distortions. In many cases the model trained using quantization is seen to be more susceptible to distortion than the selected baseline model. Moreover, there are distortions such as defocus blur where the baseline model performs better than all of the variations proposed using feature quantization \cite{FeatureQuantization}.\par

Dodge and Karam proposed the architecture of MixQualNet \cite{DodgeMixture}, an  ensemble method based on mixture of experts. Each expert in the model is trained on a particular quality degradation and the gating network predicts the type and level of distortion. Borkar and Karam proposed DeepCorrect \cite{deepcorrect} to identify and rank filters that are more susceptible to image quality reductions than other filters. An additional stack of convolutional layers are added to these filters to improve the network performance while the other filters remain unchanged. Diamond \textit{et al.} \cite{DirtyPixels} proposed an architecture that prepended a network to the classification network to produce a task-oriented intermediary image that is optimized for image classification. The model complexity and the inference speed of the above mentioned frameworks increase due to the presence of additional network layers and the prepended network, respectively. Moreover, as mentioned previously, all the aforementioned approaches are focused on the task of image classification alone and do not consider the task of object detection, which not only includes object classification at different scales but also includes object localization and bounding box regression.\par

\begin{figure*}[!t]
  \centering
  \includegraphics[width=\linewidth,height=0.315\textheight]{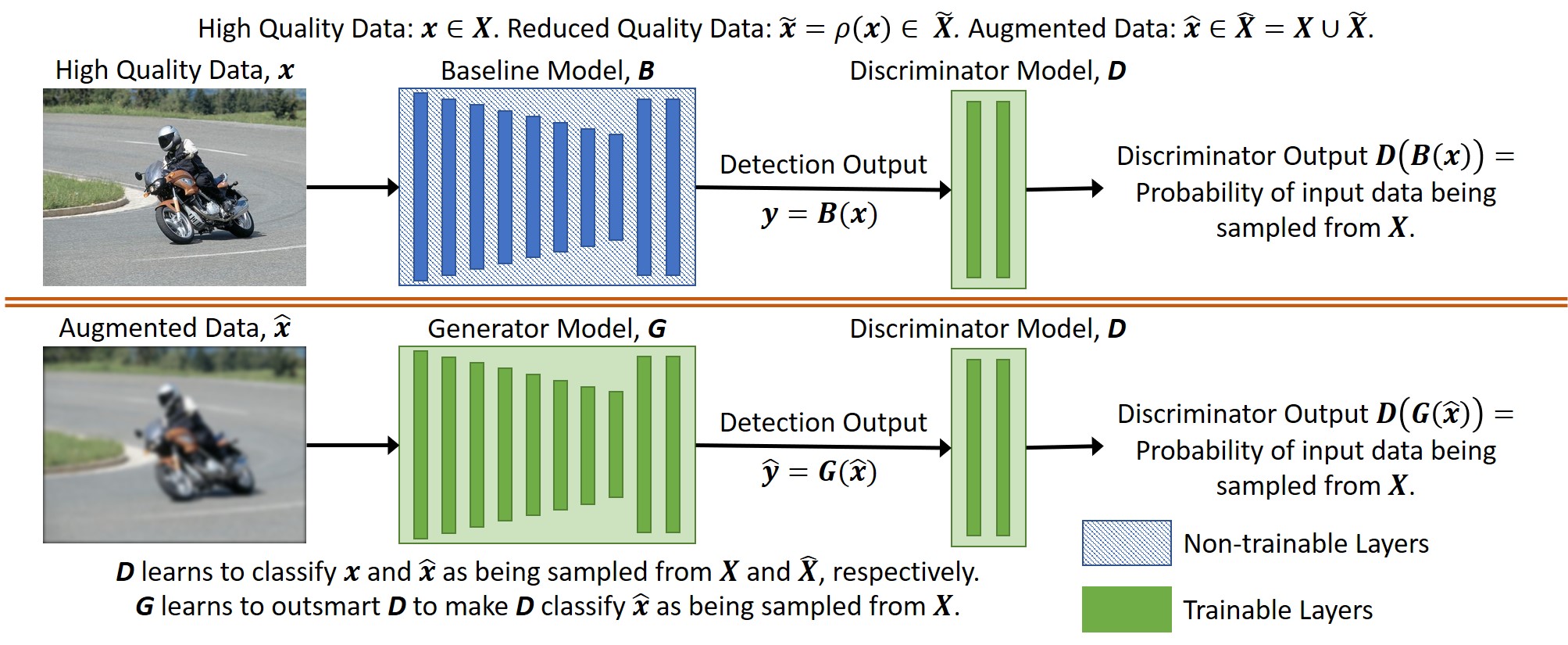}
  \caption{Block diagram of the proposed GAN-based Detection of Objects (GAN-DO) framework. The discriminator learns to distinguish between the pre-trained baseline model output \(\boldsymbol{y}\) of the high quality original data \(\boldsymbol{x}\) and the generator output \(\boldsymbol{\hat{y}}\) of the augmented data \(\boldsymbol{\hat{x}}\) with varying quality. The generator \(\boldsymbol{\textit{G}}\) learns to outsmart the discriminator \(\boldsymbol{\textit{D}}\) to make \(\boldsymbol{\textit{D}}\) classify the generator's output of the augmented data as the baseline model's output of the original data.}
  \label{BlockDia}
\end{figure*}

While there are image denoising \cite{BM3D} and image deblurring \cite{NCSR, deblurgan, GANdeblurring} methods that can be used as a pre-processing stage for object detection, these methods add significant computational overhead during inference. As an example, the work in \cite{deblurgan} requires the image to be fed-forward through 2 strided convolutions, 9 residual blocks and 2 transposed convolutional layers to generate a deblurred image. This additional computational overhead results in additional model capacity and complexity, and a latency in computing the object detection output which might not be acceptable in critical real-time applications. Moreover, recent work \cite{deepcorrect, StabilityTraining} show that denoised and deblurred images generated for improved visual perception do not translate to better performance in image understanding tasks. Additionally, one would require another blur/noise detection module to decide whether or not to use the pre-processing stage during inference in scenarios that tackle both sharp and blurred/noisy images. In the context of object detection, prior work \cite{ODdomainadapt} has shown that fine-tuning can outperform domain adaptation methods when there is availability of sufficient annotated training data. GANs have been previously used to achieve resilience in object scale variations \cite{smallObjdet} by ensuring the features for object detection of small objects look similar to the features of large objects. However, no prior work has investigated the effect of reduced quality images on object detection. Hence, there is a need to produce DNN object detection models that are robust to varying image quality.

\section{Proposed Framework}\label{ProposedFramework}
As described in the previous section, existing methods \cite{StabilityTraining} employ an additional loss term along with the loss for the intended task in order to introduce robustness to the network. However, manually choosing or designing the stability loss for each desired task and/or for each image quality level is challenging and none of the aforementioned methods propose an additional loss term for the task of object detection. Therefore, in our proposed framework, we design a network that learns an adversarial objective through training, to add robustness to the object detection model. Our proposed training framework, GAN-DO, consists of two neural networks, namely a generator and a discriminator collectively working as a GAN. However, only the generator needs to be retained during deployment of the object detection model at test time.\par

\boldmath For every high quality image \(x \in X\) in the clean original dataset \(X\), we construct the reduced quality dataset \(\widetilde{X}\), using the image quality distortion \unboldmath \(\rho_{j}(\cdot)\) \boldmath such that \(\widetilde{x}=\)\unboldmath\(\rho_{j}(\)\boldmath\(x\)\unboldmath\()\)\boldmath, where \unboldmath\(\rho_{j}(\cdot)\)\boldmath~ is a randomly picked quality reduction level from a pool of \unboldmath\(J\)\boldmath ~levels. We create an augmented dataset \(\hat{X}\) that combines the original dataset \(X\) and the reduced quality dataset \(\widetilde{X}\). This augmented dataset \(\hat{X}\) is used for training in the proposed framework. For comparison, the same augmented dataset is also used to train a fine-tuned model.\par

Fig.~\ref{BlockDia} shows the block diagram of the proposed GAN-DO framework. The framework consists of a generator and a discriminator learning to outsmart each other. During training, the object detection output \(y \in Y\) for each of the high quality images \(x\) in the training set \(X\), is computed using the pre-trained model (pre-trained on high quality images), referred henceforth as the baseline model \(B\). Similarly, the object detection output \(\hat{y} \in \hat{Y}\) for each of the varying quality images \(\hat{x}\) in the augmented dataset \(\hat{X}\) is computed using the generator \(G\). The objective of the discriminator \(D\) is to accurately classify \(y\) and \(\hat{y}\) to be originating from \(B(x)\) and \(G(\hat{x})\), respectively. The objective of the generator \(G\) is to outsmart the discriminator to make \(D\) classify \(\hat{y}\) as originating from \(B(x)\).  \par

In other words, instead of specifying an explicit stability loss term, we train the discriminator \(D\) to learn an adversarial objective that distinguishes \(B(x)\) from \(G(\hat{x})\). Based on this adversarial objective learned by \(D\), the generator \(G\) learns to minimize the distance between \(y\) from \(\hat{y}\) such that the generator's output due to \(\hat{x}\) is similar to \(y\). Upon successful completion of training, the discriminator \(D\) can be discarded and the generator \(G\) can be used instead of the baseline model to provide increased robustness to variations in image quality. The details of the loss functions, architectures and training methodologies are provided in the following subsections.\unboldmath \par

\subsection{Loss Function}\label{lossfunction}
GAN networks are known to train faster and more effectively when they are combined with the task-oriented loss \cite{pix2pix, videoprediction, GANdeblurring, ContextEncoders}. Therefore the total loss \(L_{total}\) of the proposed framework is given by:
\begin{equation}\label{Ltotal}
L_{total} = L_{OD} +\lambda L_{GAN}
\end{equation}%
where \(L_{OD}\) is the object detection loss of the baseline model and \(L_{GAN}\) is the adversarial objective of the GAN. For convenience but without loss of generality, in order to consider \(L_{total}\) as a loss function to be minimized, \(\lambda\) is a weighting factor that is set to a positive value when training the generator \(\boldsymbol{G}\) and to a negative value when training the discriminator \(\boldsymbol{D}\). The terms \(L_{OD}\) and \(L_{GAN}\) are described in more detail below.\par

The object detection loss \(L_{OD}\) is formulated as:
\begin{equation}\label{lossOD}
L_{OD} = \frac{1}{N}(L_{class} + \alpha L_{bb})
\end{equation}%
where \(N\) is the number of predicted bounding boxes, \(L_{class}\) is the classification loss, \(L_{bb}\) is the bounding box regression loss and \(\alpha\) is a hyper-parameter.
The specific choice of \(L_{class}\), \(L_{bb}\) and \(\alpha\) are model-specific loss terms and depends on the selected baseline model.\par

We consider the SSD \cite{SSD} and the RetinaNet \cite{retinanet} models for evaluating our GAN-DO framework in this paper. Both the above mentioned models employ smooth L1 loss \cite{frRCNN} for bounding box regression. SSD uses categorical cross-entropy and hard negative mining \cite{SSD} for \(L_{class}\) and \(N\) represents only a small subset of predicted bounding boxes using hard negative mining. RetinaNet uses Focal Loss \cite{retinanet} for \(L_{class}\) and uses all the predicted boxes for loss calculation. In this paper, we use the object detection loss terms as proposed for their respective models for both fine-tuning and for our GAN-DO framework, in order to provide a fair comparison.\par

\boldmath In order to make the network robust to variations in image quality, we propose to combine the object detection loss \unboldmath\(L_{OD}\)\boldmath ~with an adversarial objective \unboldmath\(L_{GAN}\)\boldmath ~during training. Consider a dataset of original data \(x \in X\) and the corresponding set of augmented data \(\hat{x} \in \hat{X}\) with probability distributions \(P_x\) and \(P_{\hat{x}}\), respectively. Let \(y \in Y\) and \(\hat{y} \in \hat{Y}\) be the corresponding outputs of \(x\) and \(\hat{x}\) computed from the baseline model \(B\) and generator \(G\), respectively, as follows (Fig. \ref{BlockDia}): \unboldmath
\begin{equation}\label{yytilde}
\begin{split}
\boldsymbol{y} = \boldsymbol{B}(\boldsymbol{x})\\
\boldsymbol{\hat{y}} = \boldsymbol{G}(\boldsymbol{\hat{x}})
\end{split}
\end{equation}\par
 
The generator \(\boldsymbol{G}\) generates an output \(\boldsymbol{G}(\boldsymbol{\hat{x}})\) using the input \(\boldsymbol{\hat{x}}\). The discriminator \(\boldsymbol{D}\) distinguishes the ``real'' original image data \(\boldsymbol{x}\) from the ``fake'' augmented image data \(\boldsymbol{\hat{x}}\) using the object detection outputs, \(\boldsymbol{B}(\boldsymbol{x})\) and \(\boldsymbol{G}(\boldsymbol{\hat{x}})\). The objective function of such  a GAN is given by: 
\begin{equation}\label{genericGANLoss}
\begin{aligned}
L_{GAN} & = \mathbb{E}_{\boldsymbol{x} \sim\boldsymbol{P_x(x)}}(\log \boldsymbol{D}(\boldsymbol{B}(\boldsymbol{x}))) \\
& ~~~ +\mathbb{E}_{\boldsymbol{\hat{x}} \sim \boldsymbol{P_{\hat{x}}(\hat{x})}}(\log(1 -  \boldsymbol{D}(\boldsymbol{G}(\boldsymbol{\hat{x}}))))
\end{aligned}
\end{equation}
where the discriminator output \(\boldsymbol{D}(\cdot)\) represents the discriminator's predicted probability of the input image belonging to the original data \(\boldsymbol{X}\). The goal of the discriminator is to maximize this objective function and the goal of the generator is to minimize this objective function. \(L_{GAN}\) is implemented as a binary cross-entropy function as discussed in~\cite{labelflip1}.\par

\subsection{Network Architecture}\label{netarch}

We consider two architectures in this paper for the task of object detection, SSD \cite{SSD} and RetinaNet \cite{retinanet}. The SSD architecture \cite{SSD} uses VGG16 \cite{VGG} as the feature extractor of the object detection network. In this paper, we use SSD300 \cite{SSD}, which corresponds to the resizing of input images to size 300 x 300 pixels. The RetinaNet architecture consists of a ResNet \cite{resnet} together with a Feature Pyramid Network (FPN) \cite{FPN} as the feature extractor. In this paper, we use RetinaNet50-400 \cite{retinanet}, which uses ResNet-50 \cite{resnet} as the feature extractor. The input images are resized such that the longest side of the image is resized to 400 pixels while maintaining the aspect ratio of the original input image. We import the baseline model architecture as the generator architecture in order to retain the model complexity and inference speed. The generator computes the object detection output in a way similar to that of the baseline model, for the input images from the augmented dataset \(\boldsymbol{\hat{X}}\).\par


 The output from the object detection models serve as input to the discriminator. The output of the discriminator is the probability of the input image being sampled from \(\boldsymbol{X}\). The discriminator of our implementation contains a single fully connected layer. In our tests we observed that increasing the model capacity of the discriminator with more layers made the discriminator so strong that the generator could not outsmart the discriminator.  Based on the size, complexity and capacity of the baseline model, the discriminator model size can be varied for other tasks. Although additional complexity is added to the framework during training by the use of the discriminator, it should be noted that the discriminator is discarded during inference. Therefore, during inference, the model complexity and speed of the model resulting from the GAN-DO framework remains identical to the baseline model. \par

\subsection{Training and Inference}\label{trainingInference}

Similar to fine-tuning, the weights for the generator in the GAN-DO framework are initialized with the pre-trained weights of the baseline model (SSD300 or RetinaNet50-400). The augmented dataset \(\boldsymbol{\hat{X}}\) is created such that it contains a uniform mix (1:1 ratio) of high quality images \(\boldsymbol{x}\) and reduced quality images \(\boldsymbol{\widetilde{x}}\) in order for the network to perform well on both high quality and reduced quality images. The pseudo-code of the proposed framework that is used to train an object detection model is given in Algorithm \ref{GAN_alg}.\par 

For each iteration of training, an image \(\boldsymbol{\hat{x}}_s\) in a random augmented mini-batch \(\boldsymbol{\hat{X}}_S\) of size \(S\) (\(s\mathtt{\sim} [1,2,...,S]\)), is selected on the fly using the equation:
\begin{equation}\label{augdataset}
\boldsymbol{\hat{x}}_s=\begin{cases}\boldsymbol{x}_s, & \text{if}~ s\leq S/2\\ \rho_{j}(\boldsymbol{x}_s), ~j \mathtt{\sim} \textit{U}[1,2,...,J],  & \text{otherwise}\end{cases}
\end{equation}
where  \(\rho_{j}\) is a randomly picked \(j\)-th quality distortion kernel from a pool of \(J\) distortion levels. Details about the distortions are presented in Section \ref{distortions}.\par

The original mini-batch  with high quality images \(\boldsymbol{x}_s\), and the augmented mini-batch with varying quality images \(\boldsymbol{\hat{x}}_s\), are used to compute the object detection outputs from the baseline and the generator models, \(\boldsymbol{B}(\boldsymbol{x}_s)\) and \(\boldsymbol{G}(\boldsymbol{\hat{x}}_s)\), respectively, as in Eq.~(\ref{yytilde}). The discriminator \(\boldsymbol{D}\) is first trained to predict \(\boldsymbol{x}_s \in \boldsymbol{X}\) (or equivalently \(\boldsymbol{y}_s \in \boldsymbol{Y}\)) and \(\boldsymbol{\hat{x}}_s \in \boldsymbol{\hat{X}}\) (or equivalently \(\boldsymbol{\hat{y}}_s \in \boldsymbol{\hat{Y}}\)), for \(\boldsymbol{B}(\boldsymbol{x}_s)\) and \(\boldsymbol{G}(\boldsymbol{\hat{x}}_s)\), respectively. The generator \(\boldsymbol{G}\) is trained to predict \(\boldsymbol{G}(\boldsymbol{\hat{x}}_s)\) such that \(\boldsymbol{D}\) predicts \(\boldsymbol{\hat{x}}_s \in \boldsymbol{X}\) (or equivalently \(\boldsymbol{\hat{y}}_s \in \boldsymbol{{Y}}\)).\par

All networks in this paper are trained using the Adam optimizer \cite{ADAM}. For fine-tuning, the decay rates of the first and second moments of gradients (\(\beta_1\) and \(\beta_2\) in \cite{ADAM}) are set to 0.9 and 0.99, respectively. In the proposed GAN framework, the decay rates of the first and second moments of gradients are set to 0.5 and 0.99, respectively, for both \(\boldsymbol{G}\) and \(\boldsymbol{D}\) as these values are shown \cite{DCGAN} to stabilize adversarial training.

\begin{algorithm}[!t]
\caption{Proposed training methodology using GAN-DO framework}
\label{GAN_alg}
{\textbf{Input:} Training dataset with original images \(\boldsymbol{X}\). Distortion kernels \(\rho_j(\cdot),~ j=1,...,J\)}. Baseline model \(\boldsymbol{B}\), Generator model \(\boldsymbol{G}\) initialized with pre-trained weights of \(\boldsymbol{B}\), Discriminator model \(\boldsymbol{D}\) with Normal initialized weights, number of training iterations \(L\) and mini-batch size~\(S\).
\begin{algorithmic}[1]
\FOR{\(l=1~to~L\)} 
\STATE {Draw a random mini-batch of images \(\boldsymbol{x}_s,~ s=1,...,S\) from training dataset \(\boldsymbol{X}\).}%
\STATE {Create the augmented mini-batch \(\boldsymbol{\hat{X}}_S\) with images \(\boldsymbol{\hat{x}}_s\) on the fly using Eq. (\ref{augdataset}).}
\STATE {Use \(\boldsymbol{B}(\boldsymbol{x}_s)\) to train \(\boldsymbol{D}\) to predict \(\boldsymbol{x}_s \in \boldsymbol{X}\). Update weights of \(\boldsymbol{D}\) to minimize Eq. (\ref{Ltotal}) with \(\lambda<0\).}
\STATE {Use \(\boldsymbol{G}(\boldsymbol{\hat{x}}_s)\) to train \(\boldsymbol{D}\) to predict \(\boldsymbol{\hat{x}}_s \in \boldsymbol{\hat{X}}\). Update weights of \(\boldsymbol{D}\) to minimize Eq. (\ref{Ltotal}) with \(\lambda<0\).}
\STATE {Train \(\boldsymbol{G}\) to predict \(\boldsymbol{G}(\boldsymbol{\hat{x}}_s)\) such that \(\boldsymbol{D}(\boldsymbol{G}(\boldsymbol{\hat{x}}_s))\)  predicts \(\boldsymbol{\hat{x}}_s \in \boldsymbol{X}\). Update weights of \(\boldsymbol{G}\) to minimize Eq. (\ref{Ltotal}) with \(\lambda>0\).}
\ENDFOR
\STATE {Discard \(\boldsymbol{D}\). Use \(\boldsymbol{G}\) to perform object detection with improved robustness to variations in image quality.}
\end{algorithmic}
\end{algorithm}\par

\section{Experimental Results}\label{results}
\subsection{Datasets and Evaluation Metrics}\label{Datasets}


Since previous work \cite{SSD} has shown that model performance can be improved by including both the PASCAL VOC2007 \cite{pascalvoc} and PASCAL VOC2012 \cite{voc2012} training images, the union of the PASCAL VOC2007 \texttt{train} and VOC2012 \texttt{trainval} images were used for training the models for both the proposed framework and fine-tuning. The PASCAL VOC2007 \texttt{val} images were used for validation (more details provided in Section \ref{GAN_gen}) and the PASCAL VOC2007 \texttt{test} images were used for testing and both Average Precision (AP) and mean Average Precision (mAP) were computed across all 20 annotated classes as per the PASCAL VOC object detection evaluation metric \cite{pascalvoc}.

\begin{figure}[!t]
  \centering
  \subfloat[]{
  \includegraphics[width=0.31\linewidth]{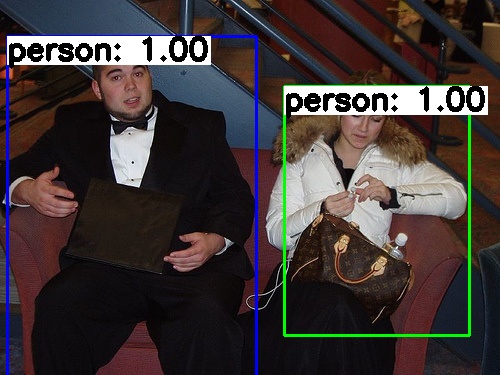}
  }
  \subfloat[]{\label{err_bb}
  \includegraphics[width=0.31\linewidth]{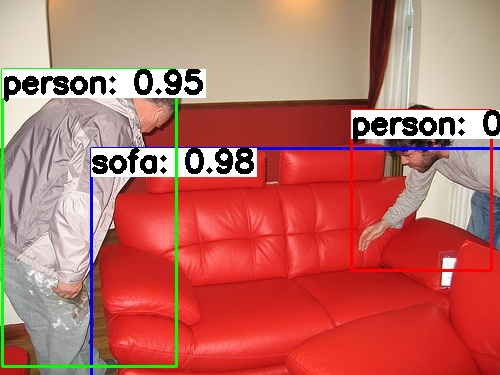}
  }
  \subfloat[]{
  \includegraphics[width=0.31\linewidth]{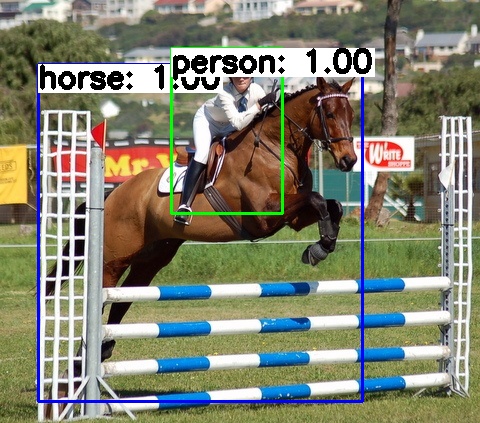}
  }
  \vfill
  \subfloat[]{\label{err_class1}
  \includegraphics[width=0.31\linewidth]{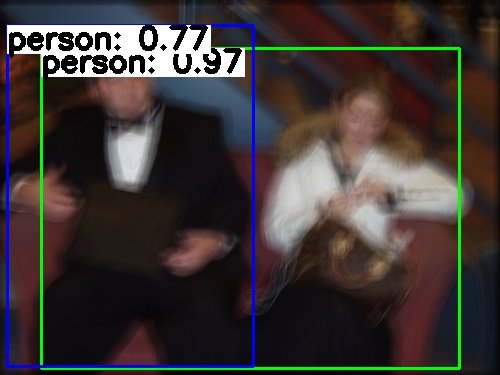}
  }
  \subfloat[]{
  \includegraphics[width=0.31\linewidth]{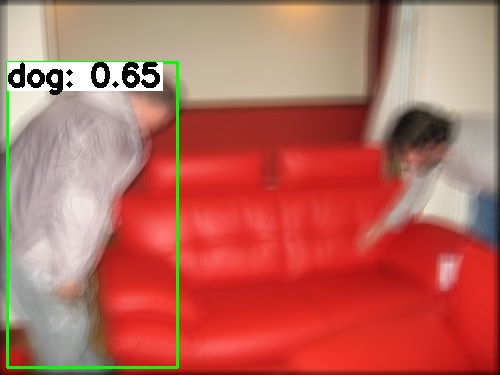}
  } 
  \subfloat[]{\label{err_class2}
  \includegraphics[width=0.31\linewidth]{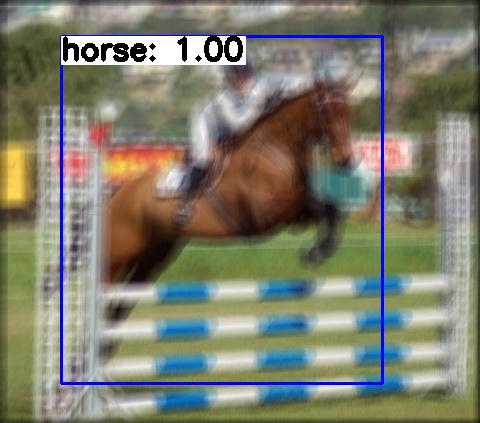}
  }
  \caption{Effect of image quality on object detection. Top row contains object detection results using the SSD300 baseline model on high quality images from the PASCAL VOC2007 \texttt{test} dataset. Bottom row contains object detection results using the SSD300 baseline model on corresponding reduced quality images (affected by camera shake blur).} 
  \label{Distortion_effect_on_OD}
\end{figure}


The Common Objects in Context (COCO) dataset \cite{coco} evaluates object detection methods over 80 object categories. All images from COCO \texttt{trainval35k} were used for training and all images from COCO \texttt{minival} were used for testing as COCO does not publicly provide labels for evaluation on their test dataset. We compute AP\textsuperscript{IoU=.50}, AP\textsuperscript{IoU=[0.50:0.05:0.95]}, AP\textsuperscript{IoU=.75}, AP\textsuperscript{small}, AP\textsuperscript{medium} and AP\textsuperscript{large} according to the COCO object detection evaluation metrics~\cite{cocometric}.\par


\subsection{Image Quality Distortions}\label{distortions}
Blur and noise are among the most commonly encountered image quality distortions in many popular applications such as social media, cellphones and autonomous driving. In this paper, we evaluate our framework using three different types of blur, namely, camera shake, Gaussian, and uniform circular average kernels referred to as defocus blurs in \cite{FeatureQuantization, BlurFinetuning}. We choose six levels of blur for Gaussian and defocus blur. The radius of the blur kernels in pixels is varied in the range [2,12] in incremental steps of 2 pixels for Gaussian blur and defocus blur. The standard deviation of each kernel of the Gaussian blur is set to one half of the radius of the blur kernel. For camera shake blur, we generate 50 random camera shake kernels as described in \cite{camshakeblur}. In this paper, we also evaluate our proposed framework on images affected by additive white Gaussian noise (AWGN). We choose five levels of AWGN with the standard deviation in the range [20,100] in incremental steps of 20. During inference, the quality of each image \(\boldsymbol{{x}}_s\) in the mini-batch was reduced by applying blur or noise of a randomly chosen level to create a reduced quality mini-batch with images \(\boldsymbol{\widetilde{x}}_s\). The models were tested on the reduced quality mini-batches for object detection accuracy.\par

\subsection{Effect of Image Quality on Bounding Box Regression and Object Classification}\label{regloss}
The effect of image quality on DNN's classification accuracy was discussed in prior work \cite{deepcorrect, DirtyPixels, dodgeUnderstanding, BlurFinetuning} for the image classification task using the ImageNet 2012 dataset (ILSVRC2012) \cite{imagenet}. The focus of this paper is on object detection (which includes both object classification and object localization) rather than the task of image classification considered in prior work. Hence, as a contribution, in addition to the object classification loss \(L_{class}\), we also investigate the effect of image quality on object bounding box regression \(L_{bb}\) (as defined in Section \ref{lossfunction}) for the object detection task. Examples of incorrect object detection on reduced quality images due to errors in bounding box regression and object classification are shown in Fig. \ref{Distortion_effect_on_OD}. Tables \ref{reglosspascal} and \ref{reglosscoco} quantify the effect of image quality on both object classification loss and object bounding box regression loss of the baseline object detection models on the PASCAL VOC2007 \texttt{test} and COCO \texttt{minival} datasets, respectively\footnote{Results on the PASCAL VOC2007 dataset are provided only for the SSD baseline model as the authors of RetinaNet \cite{retinanet} do not provide a RetinaNet baseline model trained on the PASCAL VOC dataset.}.\par

\begin{table}[!t]
\caption{Effect of image quality on object classification loss and bounding box regression loss for the SSD300 baseline model on the PASCAL VOC2007 \texttt{test} dataset. Bold numbers show best accuracy.}
\label{reglosspascal}
\centering
\resizebox{0.475\textwidth}{!}{\begin{tabular}{cccc}
\hline
Method &Distortion &Classification Loss &Regression Loss\\
\hline
\multirow{4}{*}{SSD300} &None &\textbf{1.954} &\textbf{0.973}\\
                    				 &Gaussian Blur &4.012 &1.461\\ 
                    				 &Defocus Blur &4.250 &1.544\\
                    				 &Camerashake Blur&4.305 &1.580\\
                    				 &AWGN &4.137 &1.532\\
\hline
\end{tabular}}
\end{table}

\begin{table}[!t]
\caption{Effect of image quality on object classification loss and bounding box regression loss for the baseline models with different architectures on the COCO \texttt{minival} dataset. Bold numbers show best accuracy.}
\label{reglosscoco}
\centering
\resizebox{0.475\textwidth}{!}{\begin{tabular}{cccc}
\hline
Method &Distortion &Classification Loss &Regression Loss\\
\hline
\multirow{4}{*}{SSD300} &None &\textbf{2.492} &\textbf{1.238}\\
                     &Gaussian Blur &3.347 &1.663\\ 
                     &Defocus Blur &3.518 &1.755\\
                     &Camerashake Blur &3.447 &1.732\\
                     &AWGN &3.917 &1.872\\
\hline 
\multirow{4}{*}{RetinaNet50-400} &None &\textbf{0.910} &\textbf{0.266}\\
                     &Gaussian Blur &1.272 &0.480\\ 
                     &Defocus Blur &1.364 &0.515\\
                     &Camerashake Blur &1.353 &0.471\\
                     &AWGN &1.383 &0.526\\
\hline  
												
\end{tabular}}
\end{table}

From Tables \ref{reglosspascal} and \ref{reglosscoco}, we notice that the reduced image quality results in an increase in both \(L_{class}\) and \(L_{bb}\), thereby decreasing the object detection accuracy. Camera shake blur was found to affect most the model performance for the SSD300 on the PASCAL VOC2007 \texttt{test} dataset and results in up to a 120\% increase in \(L_{class}\) and up to a 60\% increase in \(L_{bb}\), compared with the corresponding losses of the SSD300 baseline model using high quality input images. AWGN was found to affect most the model performance for both SSD300 and RetinaNet50-400 on the COCO \texttt{minival} dataset and results in up to a 57\% increase in \(L_{class}\) and 98\% increase in \(L_{bb}\). Therefore, there is a need for more robust DNN models for the task of object detection with both classification and bounding box regression features that are resilient to variations in image quality.\par

\begin{table}[t]
\caption{mAP of SSD300 based DNN models on the PASCAL VOC2007 \texttt{test} dataset using images of varying quality. Bold numbers show best accuracy.}
\label{SSDpascal}
\centering
\resizebox{0.9\linewidth}{!}{\begin{tabular}{cccc}
\hline
\multirow{2}{*}{Distortion} &\multicolumn{3}{c}{Approach}\\
&Baseline &Fine-tuning &GAN-DO\\
\hline
Gaussian Blur &44.70 &63.03 &\textbf{64.40}\\
Defocus Blur &40.77 &61.83 &\textbf{63.08}\\
Camerashake Blur &39.69 &64.63 &\textbf{67.19}\\
AWGN &42.20 &65.84 &\textbf{67.47}\\
\hline
\end{tabular}}
\end{table}

\subsection{Performance of the Proposed GAN-DO Framework}\label{GAN_gen}
For the SSD300 trained using the GAN-DO framework, the weight of the adversarial objective \(\left|\lambda\right|\) (in Eq. (\ref{Ltotal})) was set to 1. The generator was trained every iteration while the discriminator was trained every other iteration. For training SSD300 on the PASCAL VOC, the learning rate for both fine-tuning and the GAN-DO framework (generator and the discriminator) were set to 10\textsuperscript{-5}. The learning rate of all the models (fine-tuning model, generator and discriminator) was divided by 10 when the validation loss failed to improve for \(\tau\) consecutive epochs. The process of decaying the learning rate was repeated twice for the fine-tuning model and the GAN-DO framework before terminating the training process. The parameter \(\tau\) was empirically determined to provide best results when set to 4 for fine-tuning and 10 for the GAN-DO framework.\par

The performance improvement that is obtained using the GAN-DO framework on the PASCAL VOC2007 \texttt{test} dataset is shown in Table \ref{SSDpascal}. It can be seen that the GAN-DO framework results in the highest mAP as compared to the baseline and fine-tuned models across all the considered distortion types. Furthermore, the GAN-DO framework results in the highest AP across most of the object classes.\par

The ability of the GAN-DO framework to perform better at higher levels of blur is shown in Table \ref{SSDpascallevels}. Table \ref{SSDpascallevels} shows that, while the baseline model and the GAN-DO model exhibit comparable performance for the lowest blur level (blurs with 2 pixel radius), the proposed GAN-DO based model achieves the highest performance as compared to both the baseline and the fine-tuned models at all other blur levels. Table \ref{SSDpascallevels} also shows that the proposed GAN-DO based model performs better than fine-tuning on high-quality images (r=0). Therefore, training using our GAN-DO framework results in a DNN model that is more robust to varying image quality as compared to fine-tuning and the baseline model. \par

\begin{table}[!t]
\caption{mAP of SSD300 based DNN models on the PASCAL VOC2007 \texttt{test} dataset using different levels of blurs. `r' specifies the radius of the distortion kernel. Bold numbers show best accuracy.}
\label{SSDpascallevels}
\centering
\resizebox{0.47\textwidth}{!}{\begin{tabular}{ccccccccc}
\hline
Distortion &Approach &r=0 &r=2 &r=4 &r=6 &r=8 &r=10 &r=12\\
\hline
\multirow{3}{*}{Gaussian Blur} &Baseline &\textbf{77.47} &\textbf{75.67} &65.96 &51.45 &36.07 &23.84 &16.23\\
												&Fine-tuning &74.73 &73.93 &70.16 &65.47 &60.58 &55.51 &51.00\\
                                                &GAN-DO &76.17 &75.29 &\textbf{71.5} &\textbf{66.79} &\textbf{61.87} &\textbf{57.17} &\textbf{52.47}\\
 \hline
 \multirow{3}{*}{Defocus Blur} &Baseline &\textbf{77.47} &\textbf{74.68} &61.60 &44.91 &30.07 &20.17 &13.60\\
												&Fine-tuning &74.96 &73.32 &68.63 &63.98 &59.44 &54.85 &50.42\\
                                                &GAN-DO &75.94 &74.20 &\textbf{69.83} &\textbf{65.53} &\textbf{61.41} &\textbf{57.05} &\textbf{52.61}\\
\hline
\end{tabular}}
\end{table}

In order to check the robustness of the GAN-DO framework to unseen distortions, tests were conducted where the models trained on one type of blur were tested on other types of blurs. It was observed in \cite{BlurFinetuning} that fine-tuning generalizes well on unseen blur distortions for the task of image classification. However, it can be seen from Table \ref{unseen} that models trained with our GAN-DO framework perform better than fine-tuning in terms of mAP across all unseen blur distortions. Consequently, DNN models trained using our GAN-DO framework generalize better than fine-tuned models and result in an increased robustness and improved mAP accuracy on images affected by unseen distortions. \par


Table \ref{difflayers} shows the effect of the re-trainable parameters using the GAN-DO framework on the accuracy of the SSD300 based object detection. Experiments were conducted on the PASCAL VOC2007 \texttt{test} dataset with camera shake blur to compute mAPs of different configurations of the proposed framework, represented by GAN-\textit{k}. Only filters from the first layer up to the \textit{k} layer (checkpoint) were trained while the filters in the other subsequent layers remained unchanged. Instead of updating all the filters in the baseline model (represented by GAN-\textit{all\_layers}), the check point \textit{k} was set at different layers in the SSD300 architecture. The model's performance results in terms of mAP when the checkpoint \(\textit{k}\) was set at layers \texttt{pool1}, \texttt{pool2}, \texttt{pool3} and \texttt{pool4} are shown in Table \ref{difflayers}.


\begin{table}[t]
\caption{mAP comparison of SSD300 based DNN models on the PASCAL VOC2007 \texttt{test} dataset using unseen blur distortions. Bold numbers show best accuracy.}
\label{unseen}
\centering
\resizebox{0.47\textwidth}{!}{\begin{tabular}{ccccc}
\hline
\multirow{2}{*}{Trained on} &\multirow{2}{*}{Framework} &\multicolumn{3}{c}{Tested on}\\
 & &Gaussian Blur &Defocus Blur &Camerashake Blur\\\hline
\multirow{2}{*}{Gaussian Blur} &Fine-tuning &63.03 &59.97 &54.39\\ 
                          &GAN-DO	&\textbf{64.40} &\textbf{62.14} &\textbf{55.62}\\\hline
\multirow{2}{*}{Defocus Blur} &Fine-tuning &61.03 &61.83 &58.04\\ 
                          &GAN-DO	&\textbf{61.66} &\textbf{63.08} &\textbf{58.18}\\\hline
\multirow{2}{*}{Camerashake Blur} &Fine-tuning &57.28 &57.22 &64.63\\
                          &GAN-DO	&\textbf{57.91} &\textbf{59.67} &\textbf{67.19}\\\hline
\end{tabular}}
\end{table}

It can be seen from Table \ref{difflayers} that the GAN-DO framework achieves the highest mAP if the parameters are re-trained up to \texttt{pool3} (GAN-\textit{pool3}) with a comparable but slightly lower performance for GAN-\textit{pool4}. Consequently it can be observed that a higher level of robustness is achieved by re-training only the feature extractor layers instead of the entire network which includes task-specific layers for classification and localization. However, it was also observed that training all layers in the model (GAN-\textit{all\_layers}) with the proposed framework converges in fewer epochs as compared to the other configurations of GAN-\textit{k}. \par

We also test the proposed GAN-DO framework with SSD300 on a dataset larger than the PASCAL VOC such as COCO. All models (fine-tuning, generator and discriminator) were trained with a learning rate of 10\textsuperscript{-5} for the first 20 epochs and 10\textsuperscript{-6} for the next 10 epochs. Table \ref{SSDcocotable} shows the performance results that are obtained by training SSD300 with the GAN-DO framework across varying image quality. It can be seen that the GAN-DO framework results in the highest accuracy for all the types of tested quality reductions across all IoU thresholds. It can also be seen that training using the proposed framework achieves up to 12\% performance improvement in terms of AP on large objects as compared to fine-tuning. This characteristic can be highly useful in scenarios like autonomous driving where closer objects that appear larger need to be detected with higher accuracy.\par


\begin{table}[!t]
\caption{mAP comparison of SSD300 based DNN models on the PASCAL VOC2007 \texttt{test} dataset for camera shake blur. Bold numbers show best performance.}
\label{difflayers}
\centering
\resizebox{0.47\textwidth}{!}{\begin{tabular}{cccc}
\hline
Model &High Quality images &Reduced Quality images &Epochs to converge\\
\hline
Baseline &\textbf{77.47} &39.73 &-\\
Fine-tuning &74.90 &64.63 &35\\
GAN-\textit{pool1} &76.56 &56.13 &62\\
GAN-\textit{pool2} &76.51 &63.70 &58\\
GAN-\textit{pool3} &76.46 &\textbf{68.12} &66\\
GAN-\textit{pool4} &74.93 &68.08 &49\\
GAN-\textit{all\_layers} &76.12 &67.19 &\textbf{29}\\
\hline
\end{tabular}}
\end{table}

In order to show that the GAN-DO framework works on different architectures, we train RetinaNet50-400 \cite{retinanet} with the proposed framework on COCO. The weight of the adversarial objective \(\left|\lambda\right|\) (in Eq. (\ref{Ltotal})) was set to 0.5. Both the generator and the discriminator were trained at each iteration since the RetinaNet50-400 has more parameters than SSD300 and can adapt quicker based on the feedback from the discriminator. The fine-tuned model and the generator of the proposed framework were trained with a learning rate of 10\textsuperscript{-5} for the first 20 epochs and 10\textsuperscript{-6} for the next 10 epochs. The learning rate of the discriminator was set at 10\textsuperscript{-4} throughout the training process.\par

Table \ref{retinanetcocotable} shows the performance results that are obtained by training RetinaNet50-400 with the GAN-DO framework across different distortions. Compared to SSD300, RetinaNet50-400 is a larger network with more trainable parameters than SSD300. Fine-tuning benefits from the model capacity of RetinaNet50-400 to recover most of the lost accuracy in terms of AP. However, compared to the baseline and fine-tuned models, models trained using the GAN-DO framework achieve the best performance in terms of AP across all types of image quality reductions except defocus blur, while remaining comparable to fine-tuning on images affected by defocus blur. Furthermore, similar to SSD300, RetinaNet50-400 models trained using the GAN-DO framework achieve a higher AP on larger objects as compared to fine-tuning, across all types of image quality reductions.\par

\begin{table}[!t]
\caption{Object detection accuracy (AP) comparison of SSD300 based DNN models on the COCO \texttt{minival} dataset for varying image quality. Bold numbers show best accuracy.}
\label{SSDcocotable}
\centering
\resizebox{0.47\textwidth}{!}{\begin{tabular}{ccp{38pt}p{20pt}p{24pt}p{22pt}p{33pt}p{25pt}}
\hline
\multirow{2}{*}{Distortion} &\multirow{2}{*}{Method} 
&\multicolumn{3}{c}{Avg. Precision, IoU:} &\multicolumn{3}{c}{Avg. Precision, Area:}\\
& &\hfil0.50:0.95 &\hfil0.50 &\hfil0.75 &\hfil small &\hfil medium &\hfil large\\
\hline
{None} &Baseline  &\hfil\textbf{24.7} &\hfil\textbf{42.4} &\hfil\textbf{25.3} &\hfil\textbf{5.9} &\hfil\textbf{26.4} &\hfil\textbf{41.4}\\
\hline
\multirow{3}{*}{Gaussian Blur} &Baseline &\hfil14.9 &\hfil26.0 &\hfil15.1 &\hfil2.2 &\hfil13.2 &\hfil30.3\\
						  &Fine-tuning &\hfil16.6 &\hfil30.4 &\hfil16.3 &\hfil2.9 &\hfil\textbf{17.9} &\hfil31.5\\
                          &GAN-DO &\hfil\textbf{17.6} &\hfil\textbf{31.7} &\hfil\textbf{17.2} &\hfil\textbf{3.0} &\hfil17.8 &\hfil\textbf{33.3}\\
\hline
\multirow{3}{*}{Defocus Blur} &Baseline &\hfil13.8 &\hfil24.0 &\hfil13.9 &\hfil1.8 &\hfil12.1 &\hfil28.2\\
						  &Fine-tuning &\hfil15.8 &\hfil29.1 &\hfil15.4 &\hfil\textbf{2.6} &\hfil\textbf{16.8} &\hfil30.3\\
                          &GAN-DO &\hfil\textbf{17.1} &\hfil\textbf{31.0} &\hfil\textbf{17.0} &\hfil2.3 &\hfil\textbf{16.8} &\hfil\textbf{32.7}\\
\hline
\multirow{3}{*}{Camerashake Blur} &Baseline &\hfil13.6 &\hfil24.5 &\hfil13.0 &\hfil1.2 &\hfil11.6 &\hfil28.5\\
						  &Fine-tuning &\hfil16.4 &\hfil31.2 &\hfil15.5 &\hfil\textbf{2.9} &\hfil18.4 &\hfil30.7\\
                          &GAN-DO &\hfil\textbf{18.2} &\hfil\textbf{33.3} &\hfil\textbf{17.7} &\hfil2.8 &\hfil\textbf{18.6} &\hfil\textbf{33.8}\\
\hline
\multirow{3}{*}{AWGN} &Baseline &\hfil11.9 &\hfil21.3 &\hfil11.9 &\hfil2.1 &\hfil12.2 &\hfil22.6\\
						  &Fine-tuning &\hfil15.9 &\hfil30.4 &\hfil14.9 &\hfil\textbf{3.4} &\hfil18.1 &\hfil28.6\\
                           &GAN-DO &\hfil\textbf{17.8} &\hfil\textbf{32.7} &\hfil\textbf{17.3} &\hfil\textbf{3.4} &\hfil\textbf{18.6} &\hfil\textbf{32.3}\\
\hline
\end{tabular}}
\end{table}

\begin{table}[!t]
\caption{Object detection accuracy (AP) comparison of RetinaNet50-400 based DNN models on the COCO \texttt{minival} dataset for varying image quality. Bold numbers show best accuracy.}
\label{retinanetcocotable}
\centering
\resizebox{0.47\textwidth}{!}{\begin{tabular}{ccp{38pt}p{20pt}p{24pt}p{22pt}p{33pt}p{25pt}}
\hline
\multirow{2}{*}{Distortion} &\multirow{2}{*}{Method} 
&\multicolumn{3}{c}{Avg. Precision, IoU:} &\multicolumn{3}{c}{Avg. Precision, Area:}\\
& &\hfil0.50:0.95 &\hfil0.50 &\hfil0.75 &\hfil small &\hfil medium &\hfil large\\
\hline
{None} &Baseline  &\hfil\textbf{29.9} &\hfil\textbf{46.1} &\hfil\textbf{32.0} &\hfil\textbf{10.7} &\hfil\textbf{32.9} &\hfil\textbf{47.5}\\
\hline
\multirow{3}{*}{Gaussian Blur} &Baseline &\hfil16.7 &\hfil26.3 &\hfil17.6 &\hfil5.8 &\hfil16.4 &\hfil31.6\\
						  &Fine-tuning &\hfil24.1 &\hfil37.9 &\hfil25.4 &\hfil\textbf{8.3} &\hfil25.2 &\hfil40.4\\
                          &GAN-DO &\hfil\textbf{24.6} &\hfil\textbf{38.4} &\hfil\textbf{26.1} &\hfil8.0 &\hfil\textbf{25.9} &\hfil\textbf{42.1}\\
\hline
\multirow{3}{*}{Defocus Blur} &Baseline &\hfil15.1 &\hfil24.2 &\hfil15.7 &\hfil4.4 &\hfil14.3 &\hfil29.6\\
						  &Fine-tuning &\hfil\textbf{24.6} &\hfil\textbf{38.7} &\hfil\textbf{25.6} &\hfil\textbf{7.6} &\hfil\textbf{25.4} &\hfil42.2\\
                          &GAN-DO &\hfil24.5 &\hfil38.5 &\hfil\textbf{25.6} &\hfil7.1 &\hfil25.2 &\hfil\textbf{42.4}\\
\hline
\multirow{3}{*}{Camerashake Blur} &Baseline &\hfil14.6 &\hfil24.4 &\hfil14.8 &\hfil2.9 &\hfil13.5 &\hfil29.2\\
						  &Fine-tuning &\hfil24.8 &\hfil39.4 &\hfil25.7 &\hfil6.9 &\hfil26.2 &\hfil42.1\\
                          &GAN-DO &\hfil\textbf{25.4} &\hfil\textbf{40.4} &\hfil\textbf{26.4} &\hfil\textbf{7.1} &\hfil\textbf{26.8} &\hfil\textbf{43.6}\\
\hline
\multirow{3}{*}{AWGN} &Baseline &\hfil13.4 &\hfil21.7 &\hfil13.9 &\hfil4.1 &\hfil13.8 &\hfil23.0\\
						  &Fine-tuning &\hfil24.2 &\hfil38.6 &\hfil25.1 &\hfil7.5 &\hfil25.4 &\hfil40.6\\
                          &GAN-DO &\hfil\textbf{24.5} &\hfil\textbf{38.9} &\hfil\textbf{25.4} &\hfil\textbf{7.6} &\hfil\textbf{25.7} &\hfil\textbf{40.7}\\
\hline
\end{tabular}}
\end{table}

\section{Conclusion}\label{conclusion}
In this paper, we show that the accuracy of object detection networks is sensitive to images with varying quality. We propose a novel framework, called the GAN-DO framework, for re-training the parameters of the baseline model in order to increase the robustness of the model to varying image quality. The model resulting from our GAN-DO framework is identical to the baseline model in terms of model complexity and inference speed while achieving robustness to varying image quality. The GAN-DO framework outperforms the fine-tuned and baseline models across different types of tested image quality reductions and over different baseline DNN models.\par

{\small
\bibliographystyle{ieee_fullname}
\bibliography{egbib}
}

\end{document}